\documentclass[conference]{IEEEtran}
\IEEEoverridecommandlockouts
\usepackage{cite}
\usepackage{amsmath,amssymb,amsfonts}
\usepackage{algorithmic}
\usepackage{graphicx}
\usepackage{textcomp}
\usepackage{xcolor}
\def\BibTeX{{\rm B\kern-.05em{\sc i\kern-.025em b}\kern-.08em
    T\kern-.1667em\lower.7ex\hbox{E}\kern-.125emX}}
\begin{document}

\title{Pre-Avatar: An Automatic Presentation Generation Framework Leveraging Talking Avatar
}

\author{\IEEEauthorblockN{Aolan Sun\textsuperscript{+}\thanks{+ Equal contribution}, Xulong Zhang\textsuperscript{+}, Tiandong Ling\textsuperscript{+}, Jianzong Wang$^\ast$\thanks{$^\ast$Corresponding author: Jianzong Wang, jzwang@188.com.}, Ning Cheng, Jing Xiao}
\IEEEauthorblockA{\textit{Ping An Technology (Shenzhen) Co., Ltd., China}}
}

\maketitle

\begin{abstract}
Since the beginning of the COVID-19 pandemic, remote conferencing and school-teaching have become important tools. The previous applications aim to save the commuting cost with real-time interactions. However, our application is going to lower the production and reproduction costs when preparing the communication materials. This paper proposes a system called Pre-Avatar, generating a presentation video with a talking face of a target speaker with 1 front-face photo and a 3-minute voice recording. Technically, the system consists of three main modules, user experience interface (UEI), talking face module and few-shot text-to-speech (TTS) module. The system firstly clones the target speaker's voice, and then generates the speech, and finally generate an avatar with appropriate lip and head movements. Under any scenario, users only need to replace slides with different notes to generate another new video. The demo has been released here\footnote{https://pre-avatar.github.io/} and will be published as free software for use.
\end{abstract}


\section{Introduction}
\label{intro}
Since the outbreak of COVID-19, teleconferencing or distance education has become more common in real life \cite{ABUMALLOH2021101728}. Many face-to-face interactions, such as interviews, school-teaching, academic/social events, \textit{etc.}, have had to be moved online due to the pandemic \cite{10.1007/978-3-030-58858-8_32,zhang2022Singer}. Amid this trend, the workload of scholars, teachers, and business executives recording presentation videos has increased significantly. This recording effort is repetitive and not re-usable \cite{rundle2020orchestrating}. Therefore, there is a need for a system to ease these efforts, especially in the virtual space~\cite{zhang2022MetaSID}. During an online-presentation, a real-time virtual human face could shorten the 'distance' between the audience and the speaker. At the same time, it is an urgent need to generate the face with relatively small data for the cloned voice and the vivid facial expression. In this way, our technology can help greatly reduce the repetitive workload in comparison with the old-fashioned way.

\begin{figure}[th] 
    \centering 
    \includegraphics[width=\linewidth]{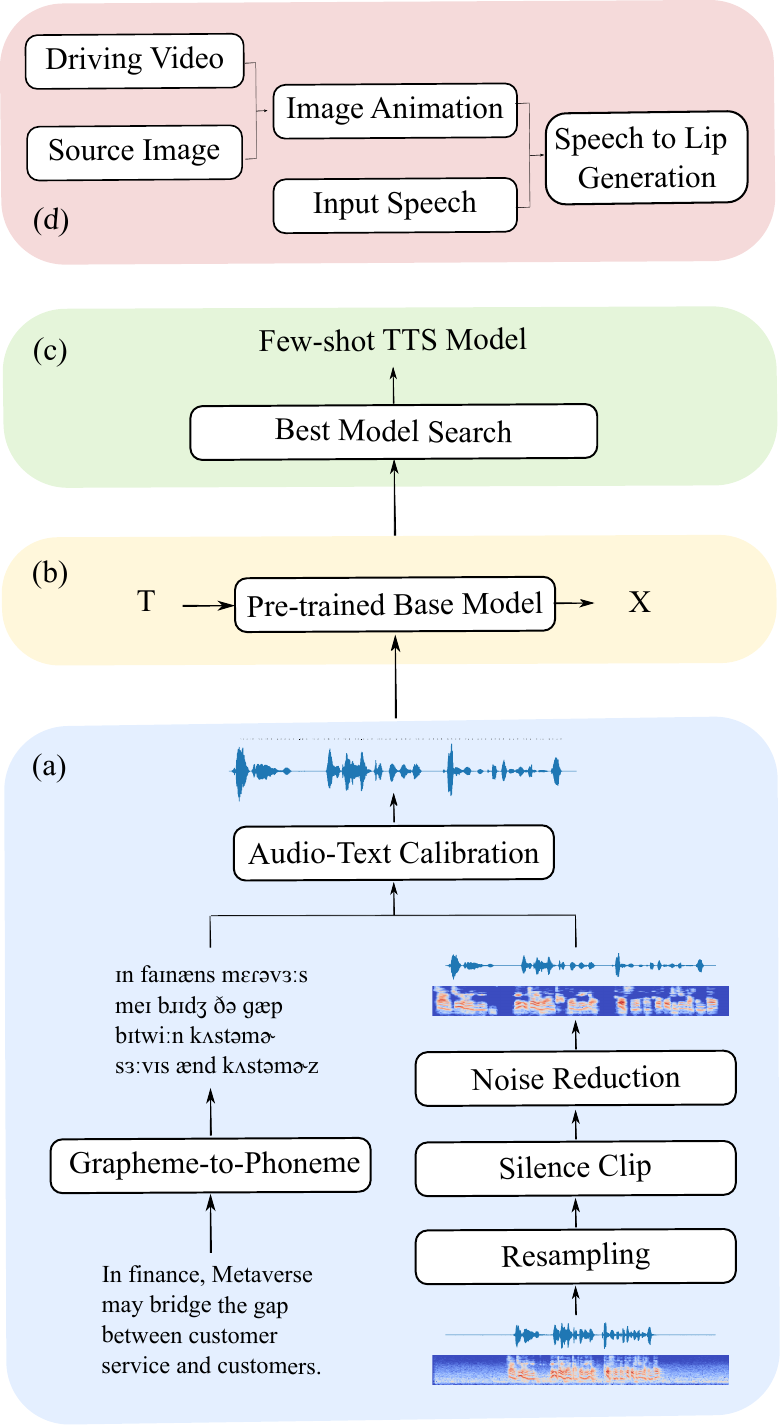}
    \caption{The architecture of the system. (a) The automatic processing stage of target speaker's audio data. (b) Model training stage of the base speaker. (c) Training and select the best fine-tuned model of the target speaker. (d) Training stage of the talking face module.}
    \label{fig:tts-train} 
\end{figure}

Since the early 2000s, some scholars have tried to provide authoring tools through which speakers can record presentation videos and upload slides \cite{tellez2007authoring}. This solves the problem of reusing presentation videos to a certain extent, but it is limited to replaying existing videos. For new topics or fields, speakers still need to spend time and effort repeatedly recording. Therefore, the combination of talking face~\cite{zhang2022shallow} and few-shot TTS~\cite{zhang2022TDASS,zhao2022nnspeech} came into being. 
The task of talking face is to generate an avatar of the target speaker using a frontal photo of the target speaker and a driving video of an arbitrary person. Methods of talking face can be divided into two categories: 3D graphics-based models and 2D-based models. The 3D graphics-based models~\cite{CaoWWSZ16} are constructed for a specific subject and animate talking face by manipulating a 3D mesh of facial models. However this methods relies on 3D face models and is difficult to generalise to arbitrary human images. More recently, a 2D-based model~\cite{Zhou000W19} leverages the power of deep generative models to generate talking faces from scratch. 
Synchronization of avatar, audio, and lip movements is also a challenging task. Prajwal \textit{et al.}~\cite{PrajwalMNJ20} proposed a method using a lip-sync discriminator to enhance lip-syncing for arbitrary talking face videos and arbitrary speech in the wild, which has achieved the best performance in generating lip movements on images or videos. 
With the development of deep models, there are some efforts to develop few-shot TTS by voice conversion~\cite{tang2022avqvc,wang2022drvc}.
But most of them can not achieve satisfactory results that can be applied in production. Furthermore, to the best of our knowledge, there is not much work on system-level multimodality of talking face and few-shot TTS.

This paper proposes a system, which consists of three modules, user experience interface (UEI), talking face and few-shot TTS module. The UEI enables users to upload materials and export outputs. The few-shot TTS can generate speech of arbitrary text based on a 3-minute recording, and the talking face module generates video of a presenter by inputting a front face photo.
Our contributions can be summarised as follows:
\begin{itemize}
\item A system is proposed to automatically generate presentation slides video by collecting data from target speakers;
\item A simple and effective few-shot TTS method is proposed to quickly clone a target speaker's voice by
3-minute audio;
\item A two-step talking face generation method is proposed
to for arbitrary images and speechs.
\end{itemize}


\section{System Architecture}
\label{sec2}
The User Experience Interface (UEI) module is displayed on the demo page and the architecture of the system in Figure \ref{fig:tts-train}. This section will introduce the training and the inference phases of the backend engine.

\subsection{System Training Stage}
\label{sec2:subsec1}
\subsubsection{Few-shot TTS}
\label{sec:subsec1:subsubsec1}
The training phase of the Few-shot TTS module is shown in Figures \ref{fig:tts-train}(a) - (c). Specifically, Figures \ref{fig:tts-train}(a) and \ref{fig:tts-train}(c) represent work on target speaker data, while the other Figure \ref{fig:tts-train}(b) shows work on the base speaker. 

Specifically, given a pre-designed text, a 3-minute recording from the presenter needs to be collected. After that, the system automatically performs five text and audio processing stages in the background, namely Resampling, Silence Clip, Noise Reduction, Grapheme-to-Phoneme and Audio-Text Calibration modules. Audio is first resampled to the target sample rate by the toolkit \textit{ffmpeg}\cite{tomar2006converting}, and then silences are clipped using a Hanning window according to a threshold, \textit{i.e.} audio below the threshold is considered silent.
This cropped speech segment is then fed into \textit{noisereduce} \cite{sainburg2020finding} to reduce ambient noise and reverberation caused by users using non-standard microphones.
At the same time the text is input into the Grapheme-to-Phoneme module and converted to the phonemic form of the International Phonetic Alphabet (IPA) via \textit{Phonemizer} \cite{zhao2022r}. Finally, Audio-Text Calibration is performed by an automatic speech recognition (ASR) system to correct mispronounced phonemes based on the audio. 

After the above 5-step process, the presenter's 3-minute paired (text, wav) data is ready. Meanwhile, the Pre-trained Base model has been prepared in the background of the system. Through transfer learning technology, the system can quickly find the learning starting point in the model training process using the processed 3-minute audio data of the presenter, and achieve rapid convergence. The final step in Few-shot TTS model training is to select the optimal model. By listening to the generated audio, it is inefficient and difficult to distinguish the quality of the audio. Therefore, the system proposes to establish quantitative criteria named Best Model Search. They are, 1) no significant high frequency noise, 2) minimize the Mel Cepstrum Distance (MCD) between the generated audio and the ground-truth audio, and 3) minimize the mel reconstruction loss value. The system displays the first five model samples that meet the three conditions to the user for final manual selection.

\subsubsection{Talking Face}
\label{sec:subsec1:subsubsec2}
The pipeline of talking face video generation is depicted in Figure~\ref{fig:tts-train}(d). First, Image Animation is trained using Driving Video and Source Image (the first frame of the Driving Video). While in the inference phase, the driving video is a human video outside the training dataset. The talking face of the presenter with the driving video action can be generated by inputting a frontal photo of the presenter. Then, Speech to Lip Generation is trained with an Image Animation model and Input Speech to adjust the mouth movements to match speech and video content.

\subsection{System Inference Stage}
\label{sec2:subsec2}
During inference, the manuscript of the presentation is fed into the Few-shot TTS model to generate the presenter's speech. At the same time, a frontal photo of the presenter with a clean background is input into the Talking Face module to generate a dynamic avatar. The generated wavs and avatars are then input into the Speech-to-lip module and aligned with the powerpoint slides, thus getting the speaker's talking video. This is portable and reusable to generate any type of presentation video.

\section{Methodology}
\label{sec3}
\subsection{Talking Face}
\label{sec3:subsec1}

\subsubsection{Image Animation}
\label{sec3:subsec1:subsubsec1}

The process of image animation is to learn the motion trajectory from a driving video, then use the image of face $S$ to animate according to the motion of the target face in the reference video $D$. Based on the FOM model, a self-supervised strategy is used for training, and a lot of videos with face category are collected. The model was used to train the reconstruction of video frames using source image and motion latent codes learned from the video frame. During training, we extracted the source image and video frames from a same driving video. The motion was encoded as the combination of local affine transformations and keypoint displacements. During inference, the trained model is applied to the image and video from different people. The static face of the source image can generate an animation with the same motion of lip and head from the reference video.

\begin{figure}
    \centering
    \includegraphics[width=\linewidth]{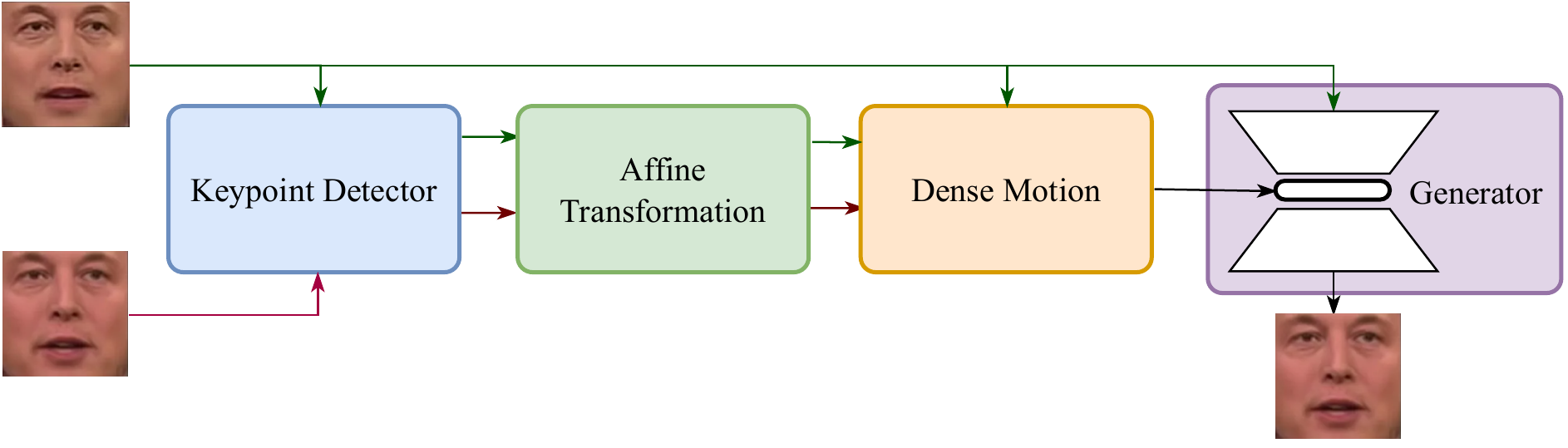}
    \caption{The pipeline of the image animation}
    \label{fig:image animation}
\end{figure}

The pipeline of Image Animation is shown in Figure~\ref{fig:image animation}. There are mainly two parts, one is the motion estimation, including Keypoint Detector, Affine Transformation, and Dense Motion, and the other is the Generator. A motion filed between reference frame $D\in \mathbb{R}^{3\times H\times W}$ in video and the source frame $S\in \mathbb{R}^{3\times H\times W}$ is predicted by the motion estimation. We can maps the keypoint location in $S$ with the point in the reference frame with the function of $T_{S\leftarrow D}$: $\mathbb{R}^2\in \mathbb{R}^2$ as the motion filed. Here backward optical flow was employed as  $T_{S\leftarrow D}$. Finally, the Generator renders the animated image based on the source image and the motion field generated by the driving frame.

\subsubsection{Speech to Lip Generation}
\label{sec3:subsec1:subsubsec2}
To align speech fragments with the talking face's lips and head movements, speech-to-lip generation is performed based on wav2lip~\cite{prajwal2020lip}.
A pre-trained expert lip-sync discriminator is used to detect synchronization in real videos. In the speech-to-lip model, consecutive video frames and audio segments are fed into the model, which is used to do the synchronization of video and audio in random windows. It contains an audio encoder and a video encoder, both built with a stack of 2D convolutional layers. The output embeddings of the video encoder and audio encoder are used to compute the L2 loss. A max-margin loss is used to contol the loss between synchronizing pairs and the unsynchronized pairs.

With the lip-sync discriminator, the generator can penalize inaccurate frames during video generation. The generator $G$ consists of three components: 1) identity encoder, 2) speech encoder, and 3) face decoder. The identity encoder encodes reference frames concatenated with the target face, where the mouth region is masked, and consists of a stack of residual convolutional layers. The speech encoder, which encodes the input speech and then concatenates it with a face representation, is constructed from a stack of 2D convolutional layers. The face decoder consists of several convolutional layers and transpose convolutional layers for upsampling. Through the minimization of reconstruction loss, the generator is trained to generate frames close to the ground-truth ones.

\subsection{Voice Clone}
\label{sec3:subsec2}
\subsubsection{Base Model}
\label{sec3:subsec2:subsubsec1}
The base model of this voice cloning module follows the end-to-end TTS framework, \textit{i.e.} VITS \cite{kim2021conditional}, where the vocoder is modified to HifiGAN-v2\cite{su2020hifigan}, and the stochastic duration predictor is pruned because the focus of this model is on voice cloning rather than generating audio with different prosody.










\subsubsection{Few-shot TTS}
\label{sec3:subsec2:subsubsec2}
Four methods including baseline models are proposed to perform the Few-shot TTS task:
\begin{itemize}
\item Benchmark: i-vector multi-speaker model. The work is following \cite{jia2018transfer} using i-vector as the speaker embedding, which can learn the voice of any speaker based on one shot data of the target speaker;
\item Imbalanced: Multi-speaker model with imbalanced data. The multi-speaker model is trained with 3 speakers, containing 2 speakers' data from sufficiently large corpus and few-shot data of the target speaker;
\item GST: the global style token (GST) model \cite{wang2018style} aims to learn representations of different styles through unsupervised learning. This means that the model can handle multi-speaker tasks if each speaker is considered to have its own style. So a GST model is first trained with multiple speakers, and then the model is fine-tuned on few-shot data of the target speaker;
\item One-to-one: Direct transfer from a single-speaker base model to a target-speaker model. The model structure remains unchanged when transfer learning is performed.
\end{itemize}

\section{Experiments and Results}
\subsection{Dataset}
During the training of talking face, there are two datasets for image animation and speech-to-lip generation respectively. The VoxCeleb \cite{nagrani2017voxceleb} dataset for learning motion from source images to target reference frames contains 22496 videos of facial objects extracted from YouTube videos. To extract face parts, a bounding box is extracted for each frame of the video. Specifically, the bounding box of the first frame is set as the initial position, and the face is tracked in the frame until it is too far from the initial position.
The smallest area which cover all the bounding boxes is set as the final crop part. This preprocessing process traverses the entire dataset video frames. Cropped sequences with resolutions lower than 256$\times$256 are filtered out. Then, the remaining cropped sequence are resized to 256$\times$256. Finally, 20047 videos are obtained, and the frames length is during 64 to 1024. The training size is set to 19522 videos and the test size is set to 525. The LRS2~\cite{afouras2018deep} dataset is used to train lip-voice synchronization, which contains news and talk videos from BBC programmes. The dataset is divided into training, validation and test set according to the broadcast date of the video. The combination of the pre-training set of 96,318 utterances and the training set of 45,839 ones is the overall training set. 1,082 utterances are used for validation and 1,242 are for testing.

For the few-shot TTS task, the datasets are LJSpeech \cite{ljspeech17}, BIAOBEI \cite{Biaobei}, and an internal Mandarin dataset, which contain 13,100, 10,000, 15,000 of 22kHz audio clips of three different female speakers. Given a pre-designed manuscript, the target speaker needs to record 3 minutes of audio.

\subsection{Experiment Setup}

For the image animation model, we set a resolution of 256$\times$256 for the dense motion predictor and the keypoint detector. For the network module, the architecture of U-Net~\cite{ronneberger2015u} is utilized with 5 conv2D blocks in the encoder, and 5 trans-conv2D blocks in the decoder, where the kernel size of the convolution layer is $(3,3)$. The network is trained using Adam optimizer with a learning rate $2e-4$ and batch size of 20. A learning decay mechanism is used at half and three-quarter training steps of the total epochs. 
For speech-to-lip generation, the model is trained with a batch size of 80 using Adam optimizer with a learning rate $1e-4$, $\beta_1=0.5$, $\beta_2=0.999$.  It should be pointed out that the weights of the lip-sync discriminator are frozen during speech-to-lip module training. 
For the few-shot TTS model,  MOSNet \cite{mosnet} is used to score the overall quality on a 5-point scale. Higher MOS scores indicate better speech quality. Similarity has the same score scale as MOS, but focuses more on audio similarity between real and synthesized audio, which is conducted crowd-sourced by 30 raters to evaluate 30 audios. Word error rate (WER) is also calculated to evaluate robustness of the model.

\subsection{Results}

\begin{figure}[thbp]
    \centering
    \includegraphics[width=\linewidth]{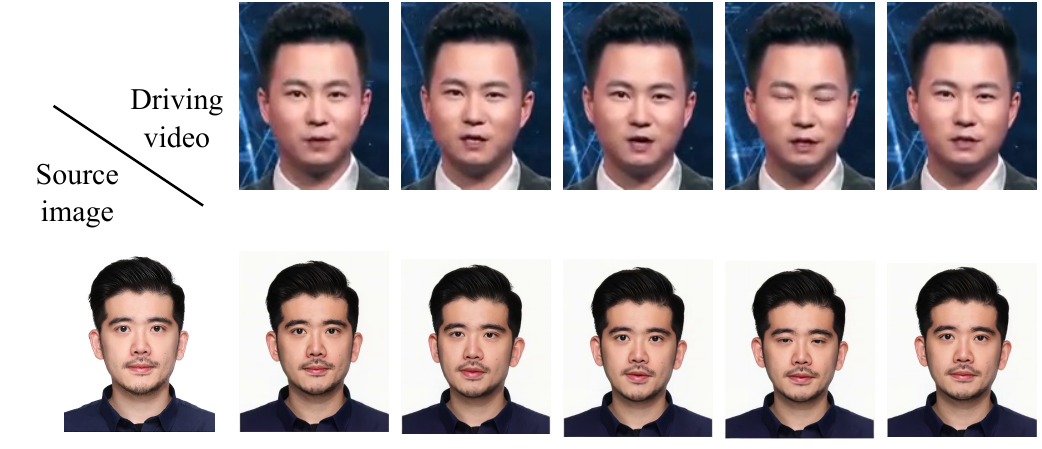}
    \caption{The results of the image animation with input a static source image to generate the target video with same motion of the driving video.}
    \label{fig: exp res image animation}
\end{figure}

\begin{figure}[thbp]
    \centering
    \includegraphics[width=\linewidth]{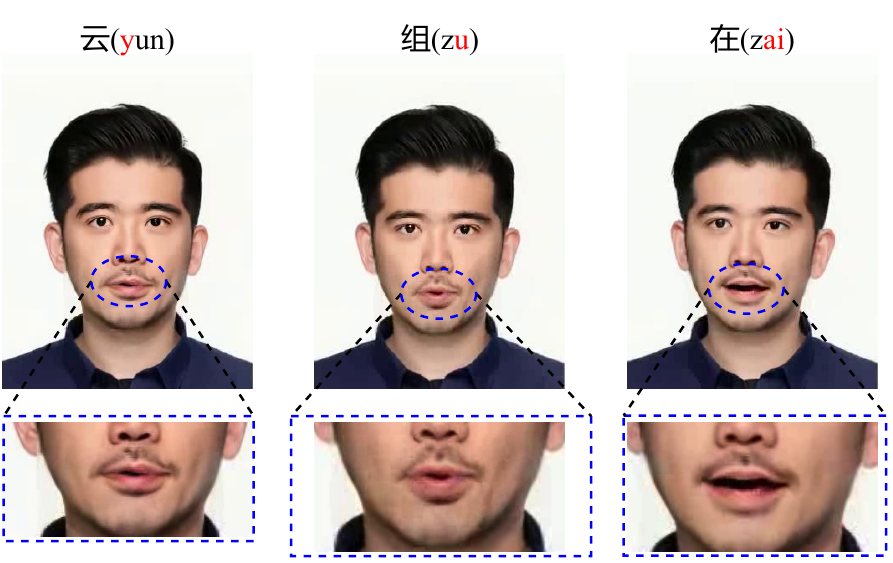}
    \caption{The result of speech to lip generation. The character was speaking  in the frame is shown in the label and with red color font. }
    \label{fig: exp res speech to lip}
\end{figure}

\begin{table}[thbp]
\caption{The results of the few-shot TTS experiments}
\begin{tabular}{cccccc}
\hline
Model               & MOS & Similarity & WER & Base & Target  \\ \hline
Benchmark           & 2.7 $\pm$ 0.02     &  4.0 $\pm$ 0.01       & 18.4\%           & 1000h  & one-shot        \\
Imbalanced          &  4.0 $\pm$ 0.03     & 3.6 $\pm$ 0.02       & 8.3\%              & 25h   & 5min       \\
GST & 3.5 $\pm$ 0.01  & 3.4 $\pm$ 0.03 & 14.3\%     & 25h & 3min       \\
One-to-one   & 4.5 $\pm$ 0.02     & 4.0 $\pm$ 0.02            & 0.9\%          & 10h  & 3min        \\ \hline
\end{tabular}
\centering

\label{tab:few-shot TTS}
\end{table}

Inputting the source image and driving video outside the training set, the results of the image animation model are shown in Figure \ref{fig: exp res image animation}. The model learns the motion trajectories of the driving video such that the static source image produces the same head and lip movements. The result of the image animation model is shown in Figure \ref{fig: exp res image animation}, with input source image and driving video. The image animation model learns the motion of the given driving video and makes the static source image vividly, and the source image and driving video both come from the training set. From the results, the pose of the source image changes significantly in columns 2 and 4, and the model can learn to detect keypoints and accurately generate target frames even in the presence of complex motion. The result generated by speech to lip is shown in Figure \ref{fig: exp res speech to lip}. The words and phonemes above the avatar are meant to show the corresponding text. As with image animation, the input speech and video are both in wild.  The results show that our model is more aucuate at lip-sync generation. We also conduct human perception tests, and let the tester to watch the synthesized video to decide the naturalness of the lip movement according to the heard audio. Specifically, our lip-synced videos are close to real synced videos through human perception tests.

As can be seen from the Table \ref{tab:few-shot TTS}, both Imbalanced and GST methods achieve reasonable results, verifying the effectiveness of both methods, but the one-to-one model achieves the highest MOS and similarity result. While benchmark only requires one-shot audio of the target speaker, the results of MOS do not achieve satisfactory results for industrial deployment. In addition, it has high requirements on the base model, requiring 1000+ hours of base audio, which is difficult to reproduce in new application scenarios. Therefore, the one-to-one method is chosen as the few-shot TTS solution for the system.


\begin{table}[thbp]
\caption{The results of system evaluation}
\begin{tabular}{ccc}
\hline
User Group  & Time Taken  & Utilization \\ \hline
Corporate executives   & 5min     &  85\%   \\
Intelligent customer servers  & 17min     & 70\%   \\
Online education teachers        & 15min  & 76\%     \\
Scientific research scholars  & 30min     & 50\%       \\ \hline
\end{tabular}
\centering

\label{tab: overall ass of PPP}
\end{table}

In addition, a comprehensive evaluation of the portability and reusability of the system is carried out. Specifically, the system provides experience for 20 people, including corporate executives, intelligent customer service, online education teachers, scientific research scholars, \textit{etc.} After they get the talking face model through the system, the system evaluates the user utilization (total hours of video generated/time spent) and the time taken by the user. The results shown in Table \ref{tab: overall ass of PPP} that it takes less than half an hour for users to collect their data so that their model can be reused in producing online presentation videos with utilization ranging from 50\% to 85\%. In theory, users can create an unlimited number of personal videos using their private keys. This demonstrates the convenience the system provides to the community.

\section{Conclusion}
In conclusion, this paper proposes a system that can generate a talking face solution of a target speaker given a frontal photo of the target speaker and a small amount of audio. When the system inputs lecture notes, the proposed system is able to automatically generate the speaker's speech video, which can help the speaker save a lot of time in repeatedly recording slideshow videos.
Previous applications have all aimed to reduce delivery costs by providing real-time and zero-commute solutions. However, when communication needs to propagate from one place to another, or even another virtual place in the Metaverse, our application will greatly reduce production and replication costs.
In addition to conference scenarios, our system can also be used in other scenarios, such as online education. 
The system is planned to be released as free software for use by the community.
\section{Acknowledgement}
This paper is supported by the Key Research and Development Program of Guangdong Province under grant No.2021B0101400003. Corresponding author is Jianzong Wang from Ping An Technology (Shenzhen) Co., Ltd (jzwang@188.com).

\bibliographystyle{IEEEtran}

\bibliography{mybib,zhangxulong}

\begin{thebibliography}{10}
\providecommand{\url}[1]{#1}
\csname url@samestyle\endcsname
\providecommand{\newblock}{\relax}
\providecommand{\bibinfo}[2]{#2}
\providecommand{\BIBentrySTDinterwordspacing}{\spaceskip=0pt\relax}
\providecommand{\BIBentryALTinterwordstretchfactor}{4}
\providecommand{\BIBentryALTinterwordspacing}{\spaceskip=\fontdimen2\font plus
\BIBentryALTinterwordstretchfactor\fontdimen3\font minus
  \fontdimen4\font\relax}
\providecommand{\BIBforeignlanguage}[2]{{%
\expandafter\ifx\csname l@#1\endcsname\relax
\typeout{** WARNING: IEEEtran.bst: No hyphenation pattern has been}%
\typeout{** loaded for the language `#1'. Using the pattern for}%
\typeout{** the default language instead.}%
\else
\language=\csname l@#1\endcsname
\fi
#2}}
\providecommand{\BIBdecl}{\relax}
\BIBdecl

\bibitem{ABUMALLOH2021101728}
R.~A. Abumalloh, S.~Asadi, M.~Nilashi, B.~Minaei-Bidgoli, F.~K. Nayer,
  S.~Samad, S.~Mohd, and O.~Ibrahim, ``The impact of coronavirus pandemic
  (covid-19) on education: The role of virtual and remote laboratories in
  education,'' \emph{Technology in Society}, vol.~67, p. 101728, 2021.

\bibitem{10.1007/978-3-030-58858-8_32}
D.~Mancl and S.~D. Fraser, ``Covid-19's influence on the future of agile,'' in
  \emph{Agile Processes in Software Engineering and Extreme Programming --
  Workshops}, M.~Paasivaara and P.~Kruchten, Eds.\hskip 1em plus 0.5em minus
  0.4em\relax Cham: Springer International Publishing, 2020, pp. 309--316.

\bibitem{zhang2022Singer}
X.~Zhang, J.~Wang, N.~Cheng, and J.~Xiao, ``Singer identification for metaverse
  with timbral and middle-level perceptual features,'' in \emph{International
  Joint Conference on Neural Networks, {IJCNN} 2022}.\hskip 1em plus 0.5em
  minus 0.4em\relax {IEEE}, 2022, pp. 178--185.

\bibitem{rundle2020orchestrating}
C.~W. Rundle, S.~S. Husayn, and R.~P. Dellavalle, ``Orchestrating a virtual
  conference amidst the covid-19 pandemic,'' \emph{Dermatology online journal},
  vol.~26, no.~7, 2020.

\bibitem{zhang2022MetaSID}
X.~Zhang, J.~Wang, N.~Cheng, and J.~Xiao, ``Metasid: Singer identification with
  domain adaptation for metaverse,'' in \emph{International Joint Conference on
  Neural Networks, {IJCNN} 2022}.\hskip 1em plus 0.5em minus 0.4em\relax
  {IEEE}, 2022, pp. 192--199.

\bibitem{tellez2007authoring}
A.~G. T{\'e}llez, ``Authoring reusable slide presentations.'' in
  \emph{Proceedings of the Third International Conference on Web Information
  Systems and Technologies}, 2007, pp. 366--371.

\bibitem{zhang2022shallow}
X.~Zhang, J.~Wang, N.~Cheng, E.~Xiao, and J.~Xiao, ``Shallow diffusion motion
  model for talking face generation from speech,'' in \emph{The 6th APWeb-WAIM
  International Joint Conference on Web and Big Data}, vol. 12858.\hskip 1em
  plus 0.5em minus 0.4em\relax Springer, 2022, pp. 300--315.

\bibitem{zhang2022TDASS}
X.~Zhang, J.~Wang, N.~Cheng, and J.~Xiao, ``Tdass: Target domain adaptation
  speech synthesis framework for multi-speaker low-resource tts,'' in
  \emph{International Joint Conference on Neural Networks, {IJCNN} 2022}.\hskip
  1em plus 0.5em minus 0.4em\relax {IEEE}, 2022, pp. 235--242.

\bibitem{zhao2022nnspeech}
B.~Zhao, X.~Zhang, J.~Wang, N.~Cheng, and J.~Xiao, ``nnspeech: Speaker-guided
  conditional variational autoencoder for zero-shot multi-speaker
  text-to-speech,'' in \emph{{IEEE} International Conference on Acoustics,
  Speech and Signal Processing, {ICASSP} 2022}.\hskip 1em plus 0.5em minus
  0.4em\relax IEEE, 2022, pp. 4293--4297.

\bibitem{CaoWWSZ16}
C.~Cao, H.~Wu, Y.~Weng, T.~Shao, and K.~Zhou, ``Real-time facial animation with
  image-based dynamic avatars,'' \emph{ACM Transactions on Graphics}, vol.~35,
  no.~4, pp. 126:1--126:12, 2016.

\bibitem{Zhou000W19}
H.~Zhou, Y.~Liu, Z.~Liu, P.~Luo, and X.~Wang, ``Talking face generation by
  adversarially disentangled audio-visual representation,'' in \emph{The
  Thirty-Third {AAAI} Conference on Artificial Intelligence}.\hskip 1em plus
  0.5em minus 0.4em\relax {AAAI} Press, 2019, pp. 9299--9306.

\bibitem{PrajwalMNJ20}
K.~R. Prajwal, R.~Mukhopadhyay, V.~P. Namboodiri, and C.~V. Jawahar, ``A lip
  sync expert is all you need for speech to lip generation in the wild,'' in
  \emph{{MM} '20: The 28th {ACM} International Conference on Multimedia}.\hskip
  1em plus 0.5em minus 0.4em\relax {ACM}, 2020, pp. 484--492.

\bibitem{tang2022avqvc}
H.~Tang, X.~Zhang, J.~Wang, N.~Cheng, and J.~Xiao, ``Avqvc: One-shot voice
  conversion by vector quantization with applying contrastive learning,'' in
  \emph{{IEEE} International Conference on Acoustics, Speech and Signal
  Processing, {ICASSP} 2022}.\hskip 1em plus 0.5em minus 0.4em\relax IEEE,
  2022, pp. 4613--4617.

\bibitem{wang2022drvc}
\BIBentryALTinterwordspacing
Q.~Wang, X.~Zhang, J.~Wang, N.~Cheng, and J.~Xiao, ``Drvc: A framework of
  any-to-any voice conversion with self-supervised learning,'' in \emph{{IEEE}
  International Conference on Acoustics, Speech and Signal Processing, {ICASSP}
  2022}.\hskip 1em plus 0.5em minus 0.4em\relax IEEE, 2022, pp. 3184--3188.
  [Online]. Available: \url{https://doi.org/10.1109/ICASSP43922.2022.9747434}
\BIBentrySTDinterwordspacing

\bibitem{tomar2006converting}
S.~Tomar, ``Converting video formats with ffmpeg,'' \emph{Linux Journal}, vol.
  2006, no. 146, p.~10, 2006.

\bibitem{sainburg2020finding}
T.~Sainburg, M.~Thielk, and T.~Q. Gentner, ``Finding, visualizing, and
  quantifying latent structure across diverse animal vocal repertoires,''
  \emph{PLoS computational biology}, vol.~16, no.~10, p. e1008228, 2020.

\bibitem{zhao2022r}
C.~Zhao, J.~Wang, X.~Qu, H.~Wang, and J.~Xiao, ``r-g2p: Evaluating and
  enhancing robustness of grapheme to phoneme conversion by controlled noise
  introducing and contextual information incorporation,'' in \emph{2022 IEEE
  International Conference on Acoustics, Speech and Signal Processing
  (ICASSP)}.\hskip 1em plus 0.5em minus 0.4em\relax IEEE, 2022, pp. 6197--6201.

\bibitem{prajwal2020lip}
K.~Prajwal, R.~Mukhopadhyay, V.~P. Namboodiri, and C.~Jawahar, ``A lip sync
  expert is all you need for speech to lip generation in the wild,'' in
  \emph{Proceedings of the 28th ACM International Conference on Multimedia},
  2020, pp. 484--492.

\bibitem{kim2021conditional}
J.~Kim, J.~Kong, and J.~Son, ``Conditional variational autoencoder with
  adversarial learning for end-to-end text-to-speech,'' in \emph{International
  Conference on Machine Learning}.\hskip 1em plus 0.5em minus 0.4em\relax PMLR,
  2021, pp. 5530--5540.

\bibitem{su2020hifigan}
J.~Su, Z.~Jin, and A.~Finkelstein, ``Hifi-gan: High-fidelity denoising and
  dereverberation based on speech deep features in adversarial networks,''
  2020.

\bibitem{jia2018transfer}
Y.~Jia, Y.~Zhang, R.~Weiss, Q.~Wang, J.~Shen, F.~Ren, P.~Nguyen, R.~Pang,
  I.~Lopez~Moreno, Y.~Wu \emph{et~al.}, ``Transfer learning from speaker
  verification to multispeaker text-to-speech synthesis,'' \emph{Advances in
  neural information processing systems}, vol.~31, 2018.

\bibitem{wang2018style}
Y.~Wang, D.~Stanton, Y.~Zhang, R.-S. Ryan, E.~Battenberg, J.~Shor, Y.~Xiao,
  Y.~Jia, F.~Ren, and R.~A. Saurous, ``Style tokens: Unsupervised style
  modeling, control and transfer in end-to-end speech synthesis,'' in
  \emph{International Conference on Machine Learning}.\hskip 1em plus 0.5em
  minus 0.4em\relax PMLR, 2018, pp. 5180--5189.

\bibitem{nagrani2017voxceleb}
A.~Nagrani, J.~S. Chung, and A.~Zisserman, ``Voxceleb: a large-scale speaker
  identification dataset,'' \emph{arXiv preprint arXiv:1706.08612}, 2017.

\bibitem{afouras2018deep}
T.~Afouras, J.~S. Chung, A.~Senior, O.~Vinyals, and A.~Zisserman, ``Deep
  audio-visual speech recognition,'' \emph{IEEE transactions on pattern
  analysis and machine intelligence}, 2018.

\bibitem{ljspeech17}
K.~Ito and L.~Johnson, ``The lj speech dataset,''
  \url{https://keithito.com/LJ-Speech-Dataset/}, 2017.

\bibitem{Biaobei}
{Databaker (Beijing) Technology Co.,Ltd.}, ``Chinese standard mandarin speech
  copus,'' https://www.data-baker.com/open\_source.html, 2020.

\bibitem{ronneberger2015u}
O.~Ronneberger, P.~Fischer, and T.~Brox, ``U-net: Convolutional networks for
  biomedical image segmentation,'' in \emph{International Conference on Medical
  image computing and computer-assisted intervention}.\hskip 1em plus 0.5em
  minus 0.4em\relax Springer, 2015, pp. 234--241.

\bibitem{mosnet}
C.-C. Lo, S.-W. Fu, W.-C. Huang, X.~Wang, J.~Yamagishi, Y.~Tsao, and H.-M.
  Wang, ``Mosnet: Deep learning based objective assessment for voice
  conversion,'' in \emph{20th Annual Conference of the International Speech
  Communication Association}, 2019.

\end{thebibliography}
\end{document}